# An Adaptive Cluster-based Filtering Framework for Speckle Reduction of OCT Skin Images

Elaheh Rashedi[1], Saba Adabi[2*], Darius Mehregan[3], Silvia Conforto[4], Xue-wen Chen[1]

[1] Department of Computer Science, Wayne State University, Detroit, Michigan, USA
[2] Postdoctoral Research Fellow at Mayo Clinic, Rochester, Minnesota, USA
[3] Department of Dermatology, Wayne State University School of Medicine, Detroit, Michigan, USA
[4] Department of Applied Electronics, Roma Tre University, Via Volterra, Rome, Italy

**Abstract:**

*Background and Objectives*

Optical coherence tomography (OCT) has become a favorable device in the dermatology discipline due to its moderate resolution and penetration depth. OCT images however contain a grainy pattern, called speckle, due to the use of a broadband source in the configuration of OCT. So far, a variety of filtering (de-speckling) techniques is introduced to reduce speckle in OCT images. Most of these methods are generic and can be applied to OCT images of different tissues. The ambition of this work is to provide a de-speckling framework specialized for filtering skin tissues for the community to utilize, adapt or build upon.

*Methods*

In this paper, we present an adaptive cluster-based filtering framework, optimized for speckle reduction of OCT skin images. In this framework, by considering the layered structure of skin, first the OCT skin images are segmented into differentiable layers utilizing clustering algorithms, and then each cluster is de-speckled individually using adaptive filtering techniques. In this study, hierarchical clustering algorithm and adaptive Wiener filtering technique are utilized to develop the framework.

*Results*

The proposed method is tested on optical solid phantoms with predetermined optical properties. The method is also tested on healthy human skin images. The results show that the proposed cluster-based filtering method can effectively reduce the speckle and increase the signal-to-noise ratio and contrast while preserving the edges in the image.

*Conclusions*

The proposed cluster-based filtering framework enable researchers to develop unsupervised learning solutions for de-speckling OCT skin images using adaptive filtering methods, or extend the framework to new applications.

**Keywords:** optical coherence tomography, skin images, speckle reduction, filtering, clustering.

---

[*] Elaheh Rashedi and Saba Adabi contributed equally to this work.

## 1. Introduction

Optical Coherent Tomography (OCT) is an optical medical imaging modality comparable to ultrasound imaging, except that OCT uses light while ultrasound uses ultrasound instead (1). OCT is utilized for performing high-resolution cross sectional imaging and works based on low-coherence interferometry (2). The interferometry relies on the temporal and spatial coherence of optical waves that are backscattered from the tissue (3). If the central wavelength of the light source is equal to or larger than the scattering compartments within the sample under investigation, the interference of the reflected light with different amplitudes and phases generates a grainy texture in the image called speckle. Speckle degrades the quality of OCT images, particularly the borders (4). By suppressing the speckle, the quality assessment measures of the images such as signal-to-noise ratio (SNR) and contrast to noise ratio (CNR) are improved and the diagnostically relevant features become more apparent. Methods for speckle reduction are divided into two main categories; hardware based methods, and software based methods (5).

The main hardware-based speckle reduction methods are compounding techniques, e.g., spatial compounding (6, 7). It has been proven that the averaging successfully reduces the noise by the factor of $\sqrt{N}$ where *N* is the number of B-scan images to be averaged if the images are sufficiently un-correlated (8). In 2012, Szkulmowski et. al. proposed a shifting beam method utilized for speckle reduction of synthetic aperture radar (SAR), ultrasound, and OCT images (5). In this method scan beams are shifted orthogonal to both light beam propagation and lateral scanning directions. The images created this way are averaged. Another compounding method is introduced by Wang et al in 2013 (9) in which the probe beam is decentered from the pivot of the scanning mirror to create multiple images that are finally averaged to obtain a single enhanced image (10-12). On the other hand, there exists software based approaches for reduction of speckle which are known as digital filtering. In 2007, Ozcan et al. discussed several digital filtering methods to decrease the speckle in OCT images (13). The authors implemented six digital filtering methods including the Enhanced Lee Filter (*ELEE*) (14), the Hybrid Median Filtering (*HMF*) (15), the Kuwahara filter, Wavelet filtering (16), methods based on artificial neural network (17-19) and the Adaptive Wiener filter (20) on OCT images. From the comparison of the obtained results, they concluded that the *ELEE* and the Wiener filter lead to an increase in the SNR and consequently to an increase in the quality of OCT images. Another filtering method is developed by Wu et al in 2015 (21), in which the statistics of the speckle such as mean and variance are measured in the OCT image and used in filtering. Total Variation (22) and Block matching and 3D filter (*BM3D*)

(23) are two edge preserving de-speckling methods. Wiener filtering method is a popular de-speckling method.

In this paper, we propose a cluster-based Filtering Framework (*CFF*) that can be utilized as a speckle noise reduction technique that utilizes the optical characteristics of skin layers' architecture. In this framework, The clustering method efficiently categorizes the areas with similar optical properties, and each region is then de-speckled using an adaptive filtering technique. In our study, hierarchical clustering algorithm and adaptive Wiener filtering technique are utilized to develop the framework.

The structure of the paper is as follows. Section 2 explains the methodology of the proposed approach, including the pseudo-code of *CFF* and its detailed explanation. The results and discussion are presented in sections 3 and 4, respectively. Finally, in section 5, the conclusion and some suggestions for future work are given.

## 2. Methodology

### 2.1. OCT system imaging

The OCT system used in this study is a multibeam, Fourier-Domain, swept-source OCT (Vivosight, Michelson Diagnostic TM Inc., Kent, UK) with a central wavelength of $1305 \pm 15$ nm. The lateral and axial resolution of our system is 7.5 µm and 10 µm, respectively. The 10 kHz sweep rate determines the time to generate one reflectivity profile. The penetration depth of the system was measured as 1.5 *mm* in healthy human skin (24).

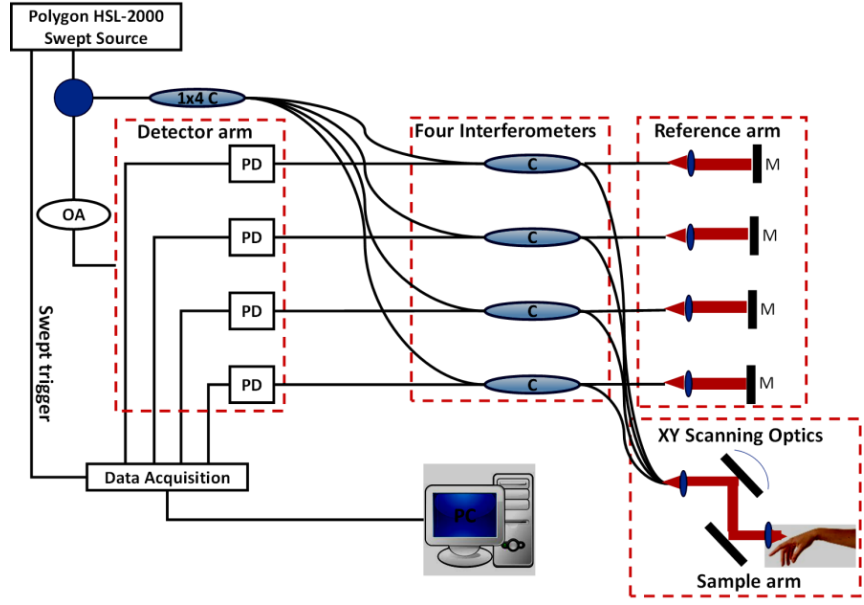

**Fig.1.** Schematic diagram of the Michelson multi-beam swept-source OCT (SS-OCT); M: mirror. C: optical coupler, M: mirror, PD: photo detector, OA: Optical attenuator (24)

The OCT is based on multi beam technology in which four 0.25 *mm* width consecutive confocal gates are combined to provide a total confocal gate of 1 *mm*. Utilizing the multi beam technology, the images obtained from the four channels are averaged. In OCT, the reflectivity profile is termed as an axial scan (A-scan or A-line). By grouping together several A-lines for different transversal positions of the incident beam on the sample, a cross section image or a B-scan is generated (25). The images obtained with this OCT system are B-Scan images with a size of 6 *mm* × 2 *mm* and software inferred C-scan images with a size of 6 *mm* × 6 *mm*. The lateral and axial resolutions of the OCT system are measured as 7.5 and 10 microns, respectively.

*2.2. Adaptive Wiener Filtering*

The Wiener, estimates the local mean and variance of a sliding window of size *n* pixel by *m* pixel, around each pixel located in the $i^{th}$ row and $j^{th}$ column of the image, and generates a new estimated pixel value of $Í_{(i,j)}$ (26). The new $Í_{(i,j)}$ is calculated as below:

$$Í_{(i,j)} = \mu + \frac{\sigma^2 - v^2}{\sigma^2}(I_{(i,j)} - \mu) \tag{1}$$

where $\mu$ is the mean, $\sigma^2$ is the variance and $v^2$ is the local variance of the sliding window of *n* by *m* pixels surrounding the pixel $(i,j)$. $v^2$ is calculated as below:

$$v^2 = \frac{1}{mn}[\frac{I^2 * O_{mn}}{mn} - (\frac{I * O_{mn}}{mn})^2] \qquad (2)$$

where $O$ is a matrix of ones with the same dimension as the sliding window, i.e. $n$ by $m$ pixels and $*$ indicating a convolution operator.

*2.3. Cluster-based Filtering Framework (CFF)*

The *CFF* algorithm begins with applying a hierarchical agglomerative clustering on each OCT image, and continues with applying and adaptive Wiener filtering on each obtained cluster. Hierarchical clustering is considered as a bottom up approach where each observation starts to create its own cluster, then pairs of clusters are merged sequentially to form one single cluster. In order to decide which cluster pairs should be combined, a measure of unlikeness between sets of observations is required (27). The measure that we used here is the Euclidean distance. The considered linkage criterion is a Ward's minimum variance method, where the objective function is the sum of squares' error (28). The pseudo code of our *CFF* is given in Algorithm. 1.

The features that we used in the clustering technique are the intensity OCT image and the map of attenuation coefficient. Attenuation coefficients are obtained by fitting a model to the OCT signal from a region of interest (ROI). Due to the scattering and absorbing structures, light is attenuated when travels within a tissue (29). Beer-Lambert law, which is governed by exponential decay, can help to explain this attenuation using the single-scattering model of the skin. Measuring attenuation coefficient from OCT images has been used in characterization of the tissue (30-33), which can consequently provide information about structural changes in the tissue. Recently, Vermeer et.al (32) developed a simple method to estimate the attenuation coefficients locally where every pixel in the OCT data set is converted into a corresponding Optical Absorption Coefficient (OAC) pixel. This produces accurate results for both homogeneous and heterogeneous tissue and does not require pre-segmenting or pre-averaging of data. The attenuation coefficient analysis method was evaluated as a diagnostic tool. The single scatter equation is determined as: $I(x) = I_0 \rho e^{-2\mu z}$, where $I$ represent the value of the detected intensity, $I_0$ is the intensity of incident light, $\rho$ is the backreflection coefficient, $\mu$ is the attenuation coefficient, and $z$ is the depth that the light is travelling through the tissue. Therefore, $z$ is the depth of penetration and can be written as a function of pixel $x$ co-ordinate, i.e. $z \propto i$. Factor of 2 comes from the fact that light travels round-trip within the tissue. The common way to calculate an attenuation coefficient is by fitting an exponential curve to the above equation (*Eq.* (3)), upon which the decay constant can be extracted. The resulted values are then averaged, smoothened, and fitted into a polynomial equation. The slope of the equation thus yields an attenuation coefficient of the region.

---

**Algorithm 1: Cluster-based Filtering Framework (*CFF*)**

**Initialization**
- $I[i,j]$      intensity of a pixel at the row of *i* and the column of *j* in the OCT image
- $\mu_a$      attenuation coefficient
- $\Delta$      pixel size
- $R$      number of pixels in each row of the OCT image
- $C$      number of pixels in each column of the OCT image
- $N$      total number of pixels ($R \times C$)
- $F$      feature vector including $\{I, \mu_a\}$
- $CL$      clustering matrix
- $K$      desired number of clusters
- $N1 \times N2$      size of neighborhood window

**1. Calculating optical properties**

**1.1.** For each pixel at the position $[i,j]$, calculate (32):
$$I[i] \propto e^{-2\mu_a z} \tag{3}$$

**2. Applying Clustering**

**2.1.** For each pixel at the position $[i,j]$, assign the feature vector of $F[i,j]$:
$$F[i,j] = \{I[i,j], \mu_a[i,j]\} \tag{4}$$

**2.2.** Apply Ward linkage hierarchical clustering on all pixels with feature set of $F$
$$CL = Hierarchical\_Linkage\,('Ward', F, K) \tag{5}$$

**2.3.** Apply max neighborhood filter on the clustering matrix $CL$ in a window size $\eta = N1 \times N2$ around each pixel
$$CL = Max\_filter\,(CL, \eta) \tag{6}$$

**3. Applying Filtering**

**3.1.** For each pixel at the position $[i,j]$ which belongs to the cluster #*k*, calculate mean and variance in a window size $\eta = N1 \times N2$
$$mean_k[i,j] = \frac{1}{\#p}\sum_{l,m \,\in\, \eta\, \&\, pixel\,[l,m]belongs\,to\,cluster\,k} I[l,m] \tag{7}$$
$$v_k^2[i,j] = \frac{1}{\#p}\sum_{l,m \,\in\, \eta\, \&\, pixel\,[l,m]belongs\,to\,cluster\,k}(I^2[l,m] - mean_k^2[i,j]) \tag{8}$$

Where **p** is the normalization constant, which is equal to the number of pixels that belongs to the cluster #*k* in the window size $\eta$.

**3.2.** For each cluster #*k*, calculate the noise variance within the cluster
$$\sigma_k^2 = \frac{1}{\#q}\sum_{pixel\,[l,m]belongs\,to\,cluster\,k} v_k^2[l,m] \tag{9}$$

Where **q** is the normalization constant, which is equal to the number of pixels that belongs to the cluster #*k*.

**3.3.** For each pixel at the position $[i,j]$ which belongs to the cluster #*k*, update the intensity values
$$I'[i,j] = mean_k[i,j] + \frac{\sigma_k^2[i,j] - v_k^2}{\sigma_k^2[i,j]}\,(I[i,j] - mean_k[i,j]) \tag{10}$$

---

The explanation of the *CFF* algorithm is given in the following.

***Initialization:*** In this step, a variable set is defined that is required to perform the *CFF* filtering method on the OCT image. The OCT image is a 2-dimentional matrix of size $R \times C$ given in terms of OCT signal intensity ($I$). Here, the desired number of clusters is set to 4, i.e., $K = 4$. The hierarchical clustering method calculates all possible clustering results with different values of *K*. Since OCT skin images can most of the time visualize four main layers of skin (Stratum corneum, epidermis, reticular dermis and papillary dermis), we color coded the results for *k=4* and our phantom is also designed for 4 layers. On the other hand, the size of window over which the algorithm estimates the local mean and variance is important. We set the size of neighborhood window to $N1 \times N2 = 5 \times 5$. As mentioned in (13, 34), the window size 5 is small enough to insure local signal stationarity, thus ensures a reasonable variance estimation.

*Calculating optical properties:* We calculate the attenuation coefficient ($\mu_a$) of each pixel at the position $[i,j]$ by using *Eq.* (3).

*Clustering:* In this step, we assign a feature vector including the intensity value ($I$) and the corresponding attenuation coefficient ($\mu_a$) of each pixel at the position $[i,j]$ (*Eq.* (4)). Thereafter, we apply the 'Ward linkage hierarchical clustering' technique on the feature vector to generate a cluster set (*Eq.* (5)). The result of the clustering is a matrix $CL$ of $R \times C$ size where each element at the position of $[i,j]$ in the matrix $CL$ indicates the cluster number that the corresponding pixel belongs to. The Elements of the matrix $CL$ are discrete values between $[1, K]$, where $K$ is the maximum number of clusters (34).

*Filtering:* After grouping the pixels into different clusters, each cluster is filtered using appropriate adaptive Wiener filter.

*2.4. Phantom design*

In order to evaluate the proposed *CFF* method, a manually segmented multilayer phantom with different optical properties is designed (6). The phantom is a virtual tissue with predefined optical properties, e.g., attenuation coefficient, scattering coefficient, and anisotropy factor. The phantom's different layers can be distinguished and labeled manually, which can be utilized later in the evaluation of the clustering algorithm. To mimic the structure of skin, each phantom has multiple layers with different optical properties. To make the solid phantom, $TiO_2$ is dissolved in polyurethane (WC-781, BJB Enterprise Co., US) (35). Different concentration of $TiO_2$ is utilized to achieve various optical properties. $TiO_2$ is dissolved into two components of polyurethane at the ratio of 100 to 85 according to the datasheet. The additives are added by 5 min vortex, followed with a 15 *min* ultrasound bath at the room temperature. Phantom is solidified overnight.

The cubic phantom has the size of 2 *cm* × 2 *cm* × 1.5 *mm*. To design each layer with the same thickness, the first and fourth layer is set to be 0.375 *mm*, where the second layer is 0.75 *mm*. For the third layer, a drop of 10 *µl* of material is added. The phantom is casted from bottom to the top by adding 150 *µl* of material on the fourth layer, 10 *µl* of material on the third layer, 280 *µl* of material on the second layer, and finally 75 *µl* for the first layer left (1L) and first layer right (1R), respectively. The schematic illustration of the phantom and the top view of the phantom are given in Fig. (2.a) and Fig. (2.b), respectively.

The Mie scattering coefficient is used to determine the reduced scattering coefficient ($\mu_s'$) of the phantom layers (36). $\mu_s'$ is calculated based on the concentration of $TiO_2$ sphere ($C_{TiO2}$) in polyurethane, which is the sphere numbers per volume of polyurethane. To calculate the $C_{TiO2}$, the volume of one sphere of $TiO_2$ ($V^S_{TiO2}$), and the volume of $TiO_2$ ($V_{TiO2}$) are calculated in *Eq.* (11) and *Eq.* (12), respectively:

$$V^S_{TiO2} = \frac{4}{3} \pi r^3 = 1.77 \times 10^{-3} \, \mu m^3 \tag{11}$$

$$V_{TiO2} = \frac{m}{\rho} = \frac{m}{4.23} \, cm^3 \tag{12}$$

Where $r$ is the radius of one $TiO_2$ sphere (diameter of $TiO_2$ is 0.15 µm); m is the total mass of $TiO_2$ in polyurethane; $\rho$ is the density of $TiO_2$, which is 4.23 $g/cm^3$.

Hence, the number of $TiO_2$ spheres ($N_{TiO2}$) is obtained by *Eq.* (13) *as*:

$$N_{TiO2} = \frac{V^S_{TiO2}}{V_{TiO2}} \tag{13}$$

Thus, the concentration of $TiO_2$ ($C_{TiO2}$), is as below (*Eq.* (14)):

$$C_{TiO2} = \frac{N_{TiO2}}{V} \tag{14}$$

where $V$ is the volume of polyurethane. Then we plug $C_{TiO2}$ into the online Mie scattering calculator (37), where we get the scattering coefficient, $\mu_s$. Finally, the reduced scattering coefficient, $\mu'_s$, can be derived from $\mu_s$, by using *Eq.* (15):

$$\mu'_s = \mu_s (1 - g) \tag{15}$$

where the value of $g$ is given as 0.715 (38, 39). Table 1 summarizes the reduced scattering coefficients of different layers of the phantom.

**Table 1.** Reduced scattering coefficient ($\mu'_s$) of different layers of the phantom (*w/v* = weight/volume, $\lambda$ = 1300 *nm*)

| | Phantom A | | | |
|---|---|---|---|---|
| | 1 | 2 | 3 | 4 |
| TiO$_2$% (*w/v*) | 0.52% | 0.26% | 0.91% | 0.65% |
| $\mu'_s$ (*cm$^{-1}$*) | 1.08 | 0.55 | 1.90 | 1.36 |

*2.5. Image quality metrics*

The definition of three common quality metrics including signal-to noise ratio (SNR), contrast to noise (CNR), edge preservation index (EPI), as well as SSIM for quantitative evaluation of de-speckling results are given in *Eq.* (16), *Eq.* (17), *Eq.* (18), and *Eq.* (19). SNR compares the signal of the OCT image to its background noise. CNR measures the contrast of the image object to background noise. EPI is a correlation measure that indicates how the edges in the image have been degraded (40, 41). And the SSIM score measures image quality based on structural similarity where two compared images, i.e. original and de-noised images, are convoluted with a circular-symmetric Gaussian weighted filter to calculate local statistics parameters (41).

$$SNR = 10 \, log_{10}(\frac{\max(I^2)}{\sigma_b^2}) \tag{16}$$

$$CNR = \frac{1}{R}(\sum_{r=1}^{R}\frac{(\mu_r - \mu_b)}{\sqrt{\sigma_r^2 + \sigma_b^2}}) \tag{17}$$

$$EPI = \frac{\sum_{i=1}^{M}\sum_{j=1}^{N}(\Delta I_{(i,j)} - \mu_{\Delta I_{(i,j)}})(\Delta \hat{I}_{(i,j)} - \mu_{\hat{I}_{(i,j)}})}{\sqrt{\sum_{i=1}^{M}\sum_{j=1}^{N}(\Delta I_{(i,j)} - \mu_{\Delta I_{(i,j)}})(\Delta \hat{I}_{(i,j)} - \mu_{\hat{I}_{(i,j)}})}} \tag{18}$$

$$SSIM = \frac{1}{MN}\sum_{i=1}^{M}\sum_{j=1}^{N}\frac{(2\mu_{I_{(i,j)}}\mu_{\hat{I}_{(i,j)}} + C_1)(2\sigma_{I_{(i,j)}}\sigma_{\hat{I}_{(i,j)}} + C_2)}{(\mu^2_{I_{(i,j)}} + \mu^2_{\hat{I}_{(i,j)}} + C_1)(\sigma^2_{I_{(i,j)}} + \sigma^2_{\hat{I}_{(i,j)}} + C_2)} \tag{19}$$

where $\mu_r$ and $\sigma_r^2$ are the mean and variance of homogenous regions of the image, $\mu_b$ and $\sigma_b^2$ are the mean and variance of the background, and $R$ is the number of ROIs which according to the literature is considered to be 10. In *Eq.* (18) and *Eq.* (19), $I$ and $\hat{I}$ are original and processed images, respectively, and $\Delta I, \Delta \hat{I}$ are the edge detected images using a standard high-pass 3×3 approximation of Laplacian operator. And $\mu_{\Delta I}, \mu_{\Delta \hat{I}}$ are the mean of images $\Delta I, \Delta \hat{I}$, respectively.

In *Eq.* (19), $C_1$ and $C_2$ are constant numbers as $C_1 = 6.5025$ and $C_2 = 58.5225$ (41). The gold standard image is generated by averaging 170 of slightly misaligned B-scan images to calculate SSIM (7).

### 3. Results

In this section, the proposed *CFF* algorithm is evaluated. The organization of this section is as follows. First, in subsection 3.1., the hierarchical clustering method is performed on the phantom to evaluate the clustering results. Then, in subsection 3.2. we apply the proposed filtering method on *in-vivo* images of human skin, and assess the results qualitatively and quantitatively.

*3.1. Evaluation of clustering method on phantom images*

To evaluate the clustering method, we imaged the phantoms described in subsection 2.4. using our OCT system.

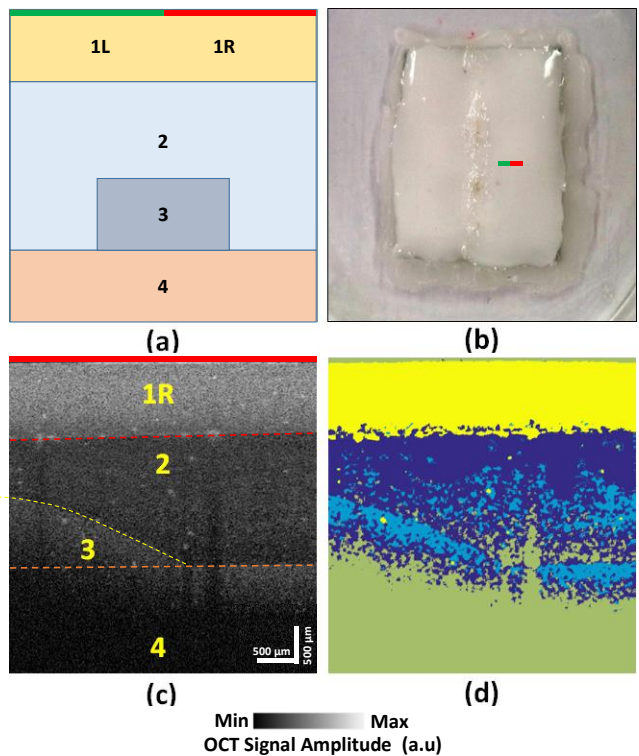

**Fig. 2.** Evaluation of clustering on OCT images of phantom. (a) Schematic illustration of a multilayer phantom, the cross-section view, (b) top view where the location of B-scan illustrated with a two-colors line, green and red, which corresponds to the two colors in the cross-section view, (c) OCT B-scan image of the right side of the phantom and manual segmentation, (d) corresponding clustered image of the OCT image.

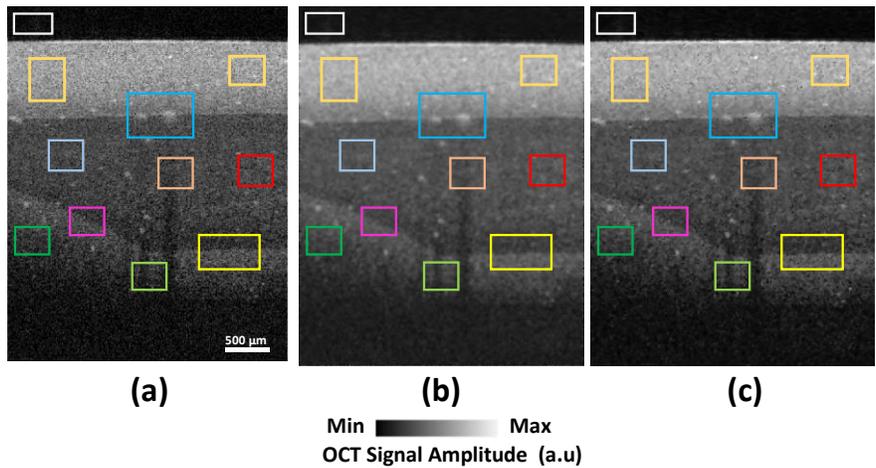

**Fig 3.** (a) Original OCT phantom image, (b) Wiener filtered phantom image, and (c) *CFF* filtered phantom image. The ROIs for calculating quality assessment measures are depicted by colored rectangles. The white rectangle on the top left of each image indicates the background noise region. Homogeneous regions are used for SNR.

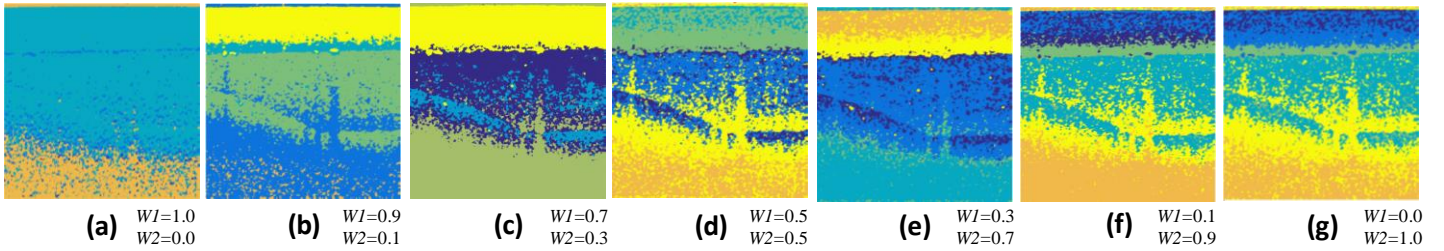

**Fig. 4.** Phantom clustering result using different combinations of features' weights, where *w1* and *w2* represent the weights of attenuation feature and intensity feature, respectively.

To evaluate the performance of the clustering method, we used phantoms with pre-defined optical properties. The OCT images of the phantom includes four distinguishable designed layers (see Fig. (2.a)). The labels related to each different layers of phantom are illustrated in Fig. (2.c). The evaluation of clustering algorithm depicting identical regions with OCT image is given in Fig. (2.d). The identical regions are presented by color coded map of the corresponding clusters, i.e. the layer 1R corresponds to the yellow cluster, the layers 2, 3 and 4 correspond to the dark blue, light blue and green clusters. Fig (2.d) show the correctness of the clustering method to differentiate layers with similar properties which is extendible to the scenario when we have complex tissues. The results of filtered phantom images using Wiener filter and *CFF* are given in Fig. (3.b) and (3.c), respectively.

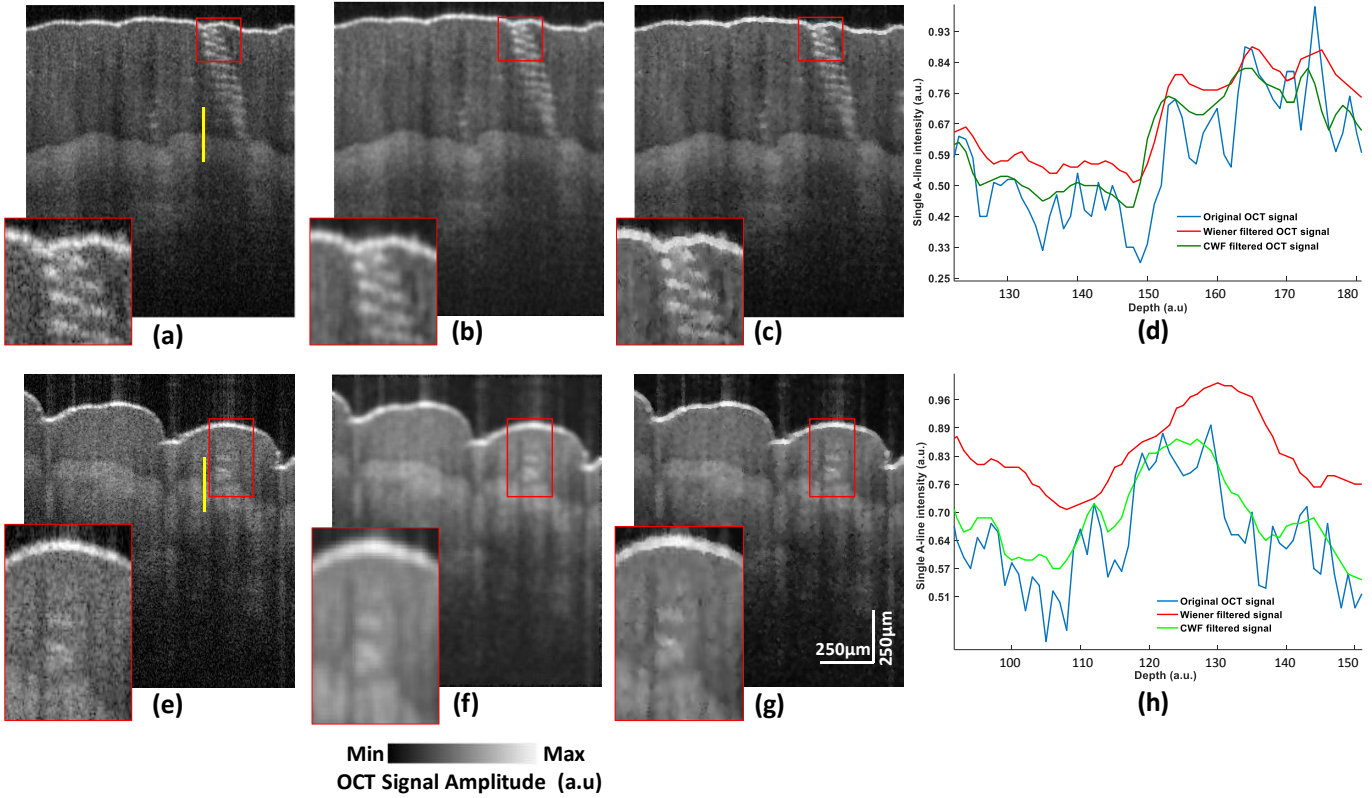

**Fig. 5.** De-speckling of OCT images. (a, e) Original OCT image, (b, f) Wiener filtered OCT image, (c, g) filtered OCT image using the proposed method, (d, h) comparison of A-line #200 profile in the original, Wiener filtered and the proposed method filtered images. Vertical yellow line corresponds to A-Line #200.

We also evaluated the clustering result on the designed phantom using different set of weights. The results are illustrated in Fig. (4), where *w1* and *w2* represent the weights of attenuation coefficient feature and intensity feature, respectively. One can claim that when the optimum weights are used in the algorithm, the effect of shadowing due to slight impurities in the phantom is reduced dramatically. As it is illustrated, the desired result was obtained using $\{w1 = 0.7, w2 = 0.3\}$ for attenuation coefficient and intensity features, respectively.

However, it is worthy to consider that due to the inhomogeneity of the $TiO_2$ particles, clustering does not differentiate the layers from each other perfectly. In addition, clustering categorizes the areas with similar speckle properties, rather than segmenting the regions from their borders.

### 3.2. Application of CFF on skin images

OCT images are acquired from 8 healthy volunteers, 25 to 35 years old male's palm of hand. The OCT machine is FDA approved. The institutional review board at Wayne State University (Independent Investigational Review Board, Detroit, MI) approved the study protocol, and informed consent was obtained from all patients before enrollment in the study. The proposed de-speckling method was applied on 170 B-scans of 8 individuals (1360 B-scans). The results obtained from *CFF* were compared with those of conventional Wiener filtering (42) both quantitatively and qualitatively (see Fig. 5). In Fig. 6, the SNR, and CNR of de-speckled images using Wiener filtering are improved by 10.4 dB, and 8.45, respectively. The improvement using the proposed method was however significantly better, which are 11.95 dB, and 10.38, respectively. The EPI was preserved better, 1.5 times, with the proposed method compared to Wiener filtering.

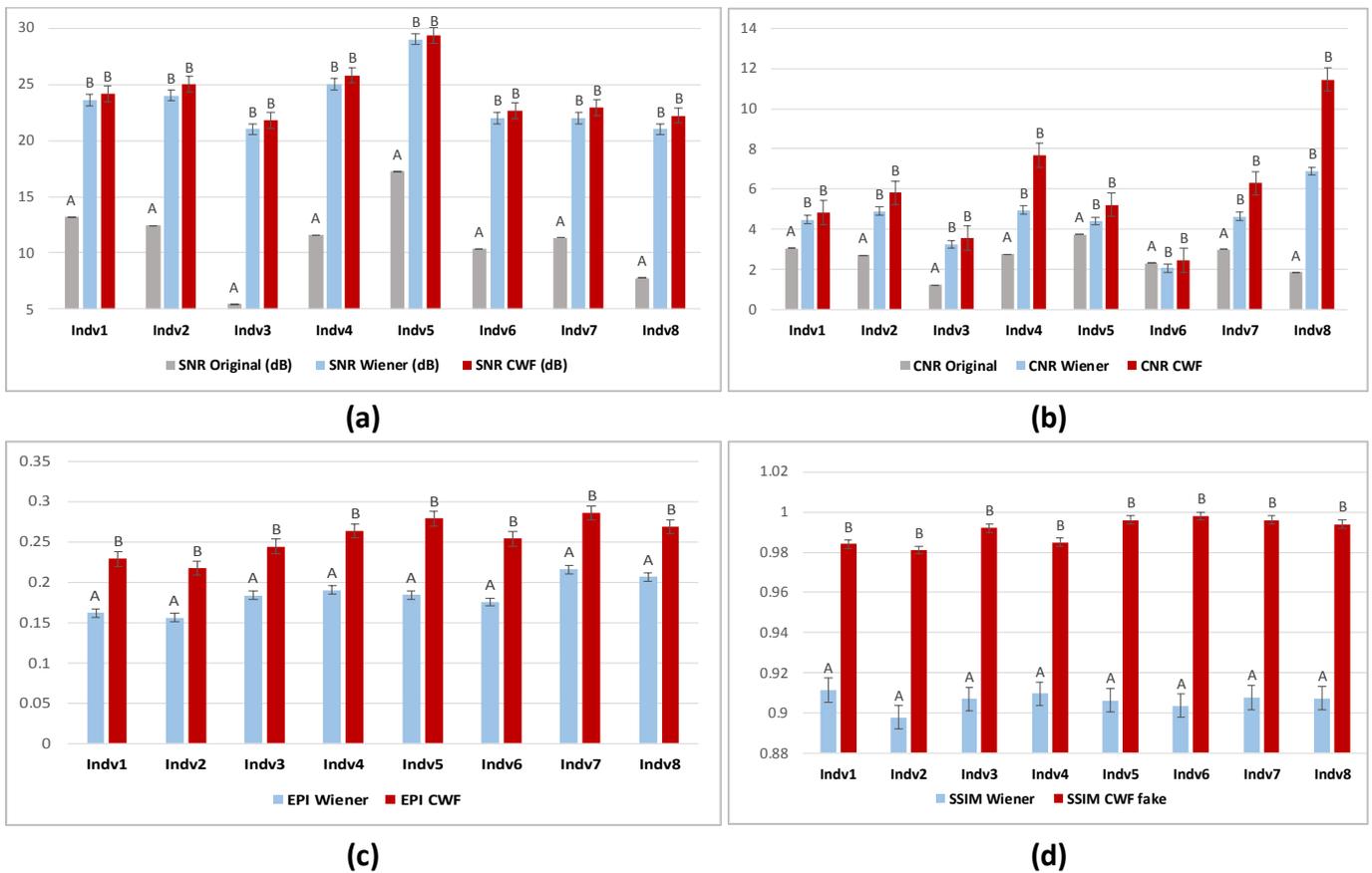

**Fig 6.** SNR, CNR, EPI, and SSIM comparison between original, Wiener filtered, and the proposed de-speckled images of 8 individuals. Bars represent standard deviation. Means with different letters (i.e., A and B) are significantly different with the *p-value* < 0.05.

The quantitative results for OCT skin images of eight individuals are listed in Fig. 6. In comparison with conventional Wiener, the *CFF* is more effective in terms of CNR. However, the SNR of filtered images by Wiener filter are slightly higher compared to those of the proposed method in some cases. The proposed filter has shown to preserve the edges better than conventional Wiener.

We also explored the relations between the optical properties of the tissue and the performance of the *CFF* algorithm. The reduced scattering coefficient ($\mu'_s$) of different layers of the phantom, and the corresponding SNR and CNR of the filtered images of the phantom are listed in Table 2. We observed that, the *CFF* algorithm performs better in the layers with lower reduced scattering coefficient ($\mu'_s$). That might be due to the smaller amount of noise in the areas with lower reduced scattering coefficients.

**Table 2.** Reduced scattering coefficient ($\mu'_s$) of different layers of the phantom, and the corresponding SNR and CNR of the *CFF* filtered phantom image

| | Phantom A | | | |
|---|---|---|---|---|
| | 1 | 2 | 3 | 4 |
| $\mu'_s$ (cm$^{-1}$) | 1.08 | 0.55 | 1.90 | 1.36 |
| SNR (dB) | 5.7311 | 12.6683 | 2.5216 | 4.8933 |
| CNR | 1.7326 | 4.9844 | 0.5161 | 0.9789 |

In comparison to Wavelet filtering (43), the *CFF* improvement of SNR, CNR, EPI, and SSIM are 2.4 (dB), 3.1, 1, and 0.08, respectively, for one given image set. The results require more exploration since the parameters in the wavelet method need to be optimized. Since the scope of this study was to improve the performance of adaptive filtering methods, we did not expand the results on such comparison.

**4. Discussion**

Although speckle is considered noise in OCT images, it carries submicron structural information of the tissue being imaged. Speckle decreases the image quality, blurs the image and conceals the diagnostically relevant features. In this study, we developed a cluster-based adaptive filtering framework that can enhance the images by considering the characteristics of the tissue in the OCT image, i.e., optical properties and intensity information. The results of the proposed method on different images showed that the de-speckled images with the proposed method are qualitatively and quantitatively improved.

Based on the results of evaluation of the proposed *CFF* method on different OCT images, we conclude that *CFF* with Wiener filtering kernel, enhances the quality of OCT images more effectively than traditional Wiener filter. Adding other statistical features of OCT images to the feature vector of our clustering method including first or higher orders statistics, is something that we will explore in the future. Other optical properties such as the scattering coefficient, anisotropy factor and geometrical properties such as shape and thickness can also be used in clustering. We also conducted a study to see the improvement of the results of *CFF* when some first statistical features are added to the feature vector, e.g., skewness and kurtosis. We observed that utilizing those parameters only adds a slight improvement to the performance of the clustering or filtering outcome.

Worth to note that the proposed framework can be utilized for boosting any other adaptive filtering method. Here, the kernel filter of the proposed cluster-based framework is Wiener filter. By replacing the filtering kernel (i.e. Wiener) with another adaptive filtering method, we can enhance the effectiveness of the filter.

*4.1. Computational performance of CFF*

Theoretically, the *CFF* algorithm implemented with Hierarchical clustering and Wiener filtering, has the order of $O((n \times m) \times log(n \times m))$ time complexity, where $n \times m$ is the size of the input image. For instance, for an input image with the size of 500×500 pixels, with a dual core processor and 4 GB memory, *CFF* takes 50 seconds to perform in comparison with Wiener filtering that takes 4 seconds.

**5. Conclusion**

In this paper, we proposed a cluster-based filtering framework for OCT skin images to reduce the speckle. The method successfully evaluated on the OCT images of tissue mimicking phantoms as well as the human skin *in-vivo*. The results show the improvement of SNR and CNR using the proposed method was significantly better than Wiener filter (11.95 dB and 10.38, vs 10.4 dB, and 8.45). The EPI was also preserved better with the proposed method compared to Wiener filtering (up to 1.5× better).

As a future work, we plan to replace our kernel filter, i.e. Wiener, with other adaptive digital filtering methods to further improve their efficiency. If a more sophisticated clustering algorithm is used, this framework helps in further functionalizing speckle reduction algorithms to improve the visibility of a specific characteristic in the OCT images.